\pdfoutput=1

\documentclass[11pt]{article}

\usepackage{naacl2021}
\usepackage{times}
\usepackage{latexsym}
\usepackage{pgfplots}
\usepackage{subfigure}
\usepackage{multirow}
\usepackage{latexsym}
\usepackage{booktabs}
\usepackage{amsmath}
\usepackage{amsfonts}
\usepackage{mathrsfs}
\usepackage{amssymb}
\usepackage{algorithm}
\usepackage{algorithmic}
\usepackage[T1]{fontenc}

\usepackage[utf8]{inputenc}

\usepackage{microtype}
\title{Self-Training for Unsupervised Neural Machine Translation\\ in Unbalanced Training Data Scenarios}

\author{ Haipeng Sun{$^{1,2}$}\thanks{\;\;Part of this work was done when Haipeng Sun and Rui Wang were an internship research fellow and a researcher at NICT, respectively.}, Rui Wang{$^3$},  Kehai Chen{$^4$}, \\ \textbf{Masao Utiyama{$^4$}, Eiichiro Sumita{$^4$}, and Tiejun Zhao{$^1$}\thanks{\;\;Corresponding author.}}\\
	$^1$Harbin Institute of Technology, Harbin, China \;\;
	$^2$JD AI Research, Beijing, China \\
	$^3$Shanghai Jiao Tong University, Shanghai, China \\
	$^4$National Institute of Information and Communications Technology (NICT), Kyoto, Japan \\
	\texttt{sunhaipeng6@jd.com},\;
	\texttt{wangrui.nlp@gmail.com},\;
	\texttt{tjzhao@hit.edu.cn} \\
	\texttt{\{khchen, mutiyama, eiichiro.sumita\}@nict.go.jp} \\
\\}

\begin{document}

\maketitle
\begin{abstract}
	Unsupervised neural machine translation (UNMT) that relies solely on massive monolingual corpora has achieved remarkable results in several translation tasks.  However, in real-world scenarios, massive monolingual corpora do not exist for some extremely low-resource languages such as Estonian, and UNMT systems usually perform poorly when  there is not adequate training corpus for one language.  In this paper, we first define and analyze the unbalanced training data scenario for UNMT. Based on this scenario, we propose UNMT self-training mechanisms to train a robust UNMT system and improve its performance in this case. Experimental results on several language pairs show that the proposed methods substantially outperform conventional UNMT systems.
\end{abstract}

\section{Introduction}
Recently, unsupervised neural machine translation (UNMT) that relies solely on massive monolingual corpora has attracted a high level of  interest in the machine translation community~\cite{DBLP:journals/corr/abs-1710-11041,lample2017unsupervised,P18-1005,lample2018phrase,wu-etal-2019-extract,sun-etal-2019-unsupervised,9043536}. With the help of cross-lingual language model pretraining~\cite{DBLP:journals/corr/abs-1901-07291,song2019mass,sun-etal-2020-knowledge},  the denoising auto-encoder~\cite{DBLP:journals/jmlr/VincentLLBM10}, and back-translation~\cite{P16-1009}, UNMT has achieved remarkable results in several translation tasks.

 However, in real-world scenarios, in contrast to the many large corpora available for high-resource languages such as English and French, massive monolingual corpora do not exist for some extremely low-resource languages such as Estonian. 
 The UNMT system usually performs poorly in a low-resource scenario when there is not an adequate training corpus for one language.   
 
 In this paper, we first define and analyze the unbalanced training data scenario for UNMT. Based on this scenario, we propose a self-training mechanism for UNMT.
 In detail, we propose self-training with unsupervised training (ST-UT) and self-training with pseudo-supervised training (ST-PT) strategies to train a robust UNMT system that performs better in this scenario.  
To the best of our knowledge, this paper is the first work to explore the unbalanced training data scenario problem in UNMT. 
Experimental results on several language pairs
show that the proposed strategies substantially outperform conventional UNMT systems.
\begin{table}[t]
	\centering
	\scalebox{0.85}{
		\begin{tabular}{lrr}
			\toprule
			
			Data size (sentences) &En-Fr & Fr-En\\
			\midrule
			50M En and 50M Fr (Baseline)& 36.63   &34.38\\
			25M En and 25M Fr& 36.59 & 34.34\\
			50M En and 2M Fr & 31.01&31.06 \\
			2M En and 50M Fr& 31.84& 30.21\\		
			2M En and 2M Fr&30.91 &29.86\\
			\bottomrule
			
	\end{tabular}}
	\caption{UNMT performance (BLEU score) for different training data sizes on En--Fr language pairs.}
	\label{tab:data_size}
\end{table}

\section{Unbalanced Training Data Scenario}

In this section, we first define the unbalanced training data scenario according to training data size. Consider one monolingual corpus $\{X\}$ in high-resource language $L_1$ and another  monolingual corpus $\{Y\}$ in low-resource language $L_2$. The data size of $\{X\}$ and  $\{Y\}$ are denoted by $|X|$ and $|Y|$, respectively. In an unbalanced training data scenario, $|X|$ is generally much larger than $|Y|$ so that training data $\{X\}$ is not fully utilized.

To investigate  UNMT performance in an unbalanced training data scenario, we empirically chose English (En) -- French (Fr) as the language pair. The detailed experimental settings for UNMT are given in Section \ref{Experiments}. We used a transformer based XLM toolkit and followed the settings of ~\citet{DBLP:journals/corr/abs-1901-07291}. We randomly extracted 2 million sentences for each language from all 50 million sentences in the En and Fr training corpora to create small corpora and simulate unbalanced training data scenarios.

Table \ref{tab:data_size} shows the UNMT performance for different training data sizes. 
The performance with 25M training sentences for both French and English configuration is similar to the baseline (50M training sentences for both French and English configuration).
However, the UNMT performance decreased substantially (4--5 BLEU points) when the size of the training data decreased rapidly. In the unbalanced training data scenario, when training data for one language was added, they were not fully utilized and only  slightly  improved the UNMT's BLEU score.  The performance (2M/50M) is similar with the UNMT system, configured 2M training sentences for both French and English. In short, Table \ref{tab:data_size} demonstrates that the UNMT performance is bounded by the smaller monolingual corpus. The UNMT model converges and even causes over-fitting in the low-resource language while the model in the high-resource language doesn’t converge.
This observation motivates us to better use the larger monolingual corpus in the unbalanced training data scenario.

\section{Background}

We first briefly describe the three components of the UNMT model~\cite{DBLP:journals/corr/abs-1901-07291}: cross-lingual language model pre-training, the denoising auto-encoder~\cite{DBLP:journals/jmlr/VincentLLBM10}, and back-translation~\cite{P16-1009}. Cross-lingual language model pre-training provides a naive bilingual signal that enables the back-translation   to   generate   pseudo-parallel corpora at the beginning of the training. 
The denoising auto-encoder acts as a language model to improve translation quality by randomly performing local substitutions and  word reorderings.  

Generally, back-translation plays an important role in achieving unsupervised translation across two languages. The pseudo-parallel sentence pairs produced by the model at the previous iteration are used to train the new translation model. The general back-translation probability is optimized by maximizing

{\footnotesize
\begin{equation}
\begin{aligned}  
\mathcal{L}_{bt} &= \mathbb{E}_{X\sim P(X)}\mathbb{E}_{Y\sim P_{M^{U^*}}(Y|X)} log P_{M^U}(X|Y)\\&+\mathbb{E}_{Y\sim P(Y)}\mathbb{E}_{X\sim P_{M^{U^*}}(X|Y)} log P_{M^U}(Y|X),
\end{aligned}
\end{equation}}%
where $P(X)$ and $P(Y)$ are the empirical data distribution from monolingual corpora $\{X\},\{Y\}$, 
and $P_{M^{U}}(Y|X)$ and $P_{M^{U}}(X|Y)$ are the conditional distributions generated by the UNMT model.
In addition, $M^{U^*}$ denotes the model at the previous iteration for generating new pseudo-parallel sentence pairs to update the UNMT model.

Self-training proposed by \citet{DBLP:journals/tit/Scudder65a}, is a semi-supervised approach that utilizes unannotated data to create better models. Self-training has been successfully applied to many natural language processing tasks~\cite{yarowsky-1995-unsupervised,mcclosky-etal-2006-effective,zhang-zong-2016-exploiting,he2019revisiting}. Recently, \citet{he2019revisiting} empirically found that noisy self-training could improve the performance of supervised machine translation and synthetic data could play a positive role, even as a target.
\section{Self-training Mechanism for UNMT}
Based on these previous empirical findings and analyses, we propose a self-training  mechanism to generate synthetic training data for UNMT to alleviate poor performance in the unbalanced training data scenario.
 The synthetic data increases the diversity of low-resource language  data, further enhancing the performance of the translation, even though the synthetic data may be noisy. As the UNMT model is trained, the quality of synthetic  data  becomes better, causing less and less noise.
Compared with the original UNMT model that the synthetic data is just used as the source part, we also use the synthetic data as the target part in our proposed methods.  Newly generated synthetic data, together with original monolingual data, are fully utilized to train a robust UNMT system in this scenario.  
According to the usage of the generated synthetic training data, our approach can be divided into two strategies: ST-UT (Algorithm \ref{alg:A}) and ST-PT (Algorithm \ref{alg:B}).


\begin{algorithm}[htb]
	\caption{ST-UT strategy}
	\label{alg:A}
	\begin{algorithmic}[1]
	\REQUIRE  
Monolingual training data $\{X\}$, $\{Y\}$\\
		\STATE {Train a UNMT model $M_{0}^U$  on monolingual training data $\{X\}$, $\{Y\}$}  
		\WHILE {epoch $l$ $\leq$ max epoch $k_1$ }
	\STATE Select a subset$\{X^{sub}\}$ randomly on monolingual training data $\{X\}$	
		\STATE Apply the last trained UNMT model $M_{l-1}^U$ to this subset$\{X^{sub}\}$ to generate synthetic data $\{Y_{M}^{sub}\}=\{M_{l-1}^U(X^{sub})\}$
		
		\STATE Train a new UNMT model $M_{l}^U$on monolingual data $\{X\}$, $\{Y\}$ and synthetic data $\{Y_{M}^{sub}\}$
		\ENDWHILE 
		\ENSURE 
		The final translation model $M_{k_1}^U$\\
	\end{algorithmic}
\end{algorithm}

\textbf{ST-UT:} 
In this strategy, we first train a UNMT model on the existing monolingual training data. The final UNMT system is trained using the ST-UT strategy for $k_1$ epochs.
For one epoch $l$ in the ST-UT strategy,  a subset$\{X^{sub}\}$ is selected randomly from monolingual training data $\{X\}$. The quantity of $\{X^{sub}\}$ is $\epsilon$ of $|X|$, $\epsilon$ is a quantity ratio hyper-parameter.
The last trained UNMT model $M_{l-1}^U$ is used  to generate synthetic data $\{Y_{M}^{sub}\}=\{M_{l-1}^U(X^{sub})\}$.
The synthetic data are used \footnote{ In contrast to using all synthetic data, we tried to train a language model and select more fluent synthetic data according to a language model perplexity score. This did not improve translation performance.}, together with the monolingual data to 
train a new UNMT model $M_{l}^U$. Therefore, the translation probability  for the ST-UT strategy is optimized by maximizing

{\footnotesize
\begin{equation}
\begin{aligned}  
\mathcal{L}_{bt} &= \mathbb{E}_{X\sim P(X)}\mathbb{E}_{Y\sim P_{M^{U^*}_l}(Y|X)} log P_{M^U_l}(X|Y)\\&+\mathbb{E}_{Y\sim P(Y)}\mathbb{E}_{X\sim P_{M^{U^*}_l}(X|Y)} log P_{M^U_l}(Y|X)
\\&+\mathbb{E}_{Y\sim P_{M^{U^*}_{l-1}}(Y|X)}\mathbb{E}_{X\sim P_{M^{U^*}_l}(X|Y)} log P_{M^U_l}(Y|X),
\end{aligned}
\end{equation}}%
where 
$P_{M^{U}_l}(Y|X)$ and $P_{M^{U}_l}(X|Y)$ are the conditional distribution generated by the UNMT model on epoch $l$ for the ST-UT strategy and $P_{M^{U^*}_{l-1}}(Y|X)$ is the conditional distribution generated by the  UNMT model on epoch $l-1$ for the ST-UT strategy.

\textbf{ST-PT:} In this strategy, we first train a UNMT system on the existing monolingual training data and switch to a standard neural machine translation system from UNMT system with synthetic parallel data for both translation directions.
The final translation system is trained using the ST-PT strategy  for $k_2$ epochs.
For one epoch $q$ in the ST-PT strategy,  a subset$\{X^{sub}\}$ is selected randomly from monolingual training data $\{X\}$,  and all monolingual data $\{Y\}$ is selected.	The quantity of $\{X^{sub}\}$ is $\epsilon$ of $|X|$, $\epsilon$ is a quantity ratio hyper-parameter.
The last trained pseudo-supervised neural machine translation (PNMT) model\footnote{Only synthetic parallel data were used to train PNMT model. } $M_{q-1}^P$ is used  to generate $\{Y_{M}^{sub}\}=\{M_{q-1}^P(X^{sub})\}$ and $\{X_{M}^{all}\}=\{M_{q-1}^P(Y^{all})\}$ to create synthetic parallel data $\{X^{sub},Y_{M}^{sub}\}$ and $\{Y^{all},X_{M}^{all}\}$. Note that we use the  UNMT model to generate synthetic parallel data during the first epoch of the ST-PT strategy.
Synthetic parallel data $\{X^{sub},Y_{M}^{sub}\}$ and $\{Y^{sub},X_{M}^{sub}\}$ are selected to 
train a new PNMT model $M_{q}^P$ that can generate translation in both directions. Therefore, the translation probability for ST-PT strategy is optimized by maximizing
\begin{algorithm}[t]
	\caption{ST-PT strategy}
	\label{alg:B}
\begin{algorithmic}[1]
\REQUIRE  
Monolingual training data $\{X\}$, $\{Y\}$\\
\STATE {Train a UNMT model $M_{0}^U$  on monolingual training data $\{X\}$, $\{Y\}$}  
\WHILE {epoch $q$ $\leq$ max epoch $k_2$ }
		\STATE Select a subset$\{X^{sub}\}$ randomly on monolingual training data $\{X\}$ and all monolingual training data $\{Y^{all}\}$
\STATE Apply  the last trained PNMT model $M_{q-1}^P (M_0^P = M_{0}^U)$ to generate $\{Y_{M}^{sub}\}=\{M_{q-1}^P(X^{sub})\}$ and $\{X_{M}^{all}\}=\{M_{q-1}^P(Y^{all})\}$
\STATE Train a new PNMT model $M_{q}^P$ on synthetic parallel corpora $\{X^{sub},Y_{M}^{sub}\}$ and $\{Y^{all},X_{M}^{all}\}$
		\ENDWHILE  
		\ENSURE 
		The final translation model $M_{k_2}^P$\\		
	\end{algorithmic}
\end{algorithm}

{\footnotesize
	\begin{equation}
	\begin{aligned}  
	\mathcal{L}_{bt} &= \mathbb{E}_{X\sim P(X)}\mathbb{E}_{Y\sim P_{M^{P^*}_{q-1}}(Y|X)}log P_{M^P_q}(X|Y)
	\\&+ \mathbb{E}_{X\sim P(X)}\mathbb{E}_{Y\sim P_{M^{P^*}_{q-1}}(Y|X)}log P_{M^P_q}(Y|X) \\&+\mathbb{E}_{Y\sim P(Y)}\mathbb{E}_{X\sim P_{M^{P^*}_{q-1}}(X|Y)} log P_{M^P_q}(Y|X)\\&+\mathbb{E}_{Y\sim P(Y)}\mathbb{E}_{X\sim P_{M^{P^*}_{q-1}}(X|Y)}log P_{M^P_q}(X|Y),
	\end{aligned}
	\end{equation}}%
where 
$P_{M^{P}_q}(Y|X)$ and $P_{M^{P}_q}(X|Y)$ are the conditional distributions generated by the PNMT model on epoch $q$ for the ST-PT strategy; $P_{M^{P^*}_{q-1}}(Y|X)$ and $P_{M^{P^*}_{q-1}}(X|Y)$ are the conditional distributions generated by the PNMT model on epoch $q-1$ for the ST-PT strategy.

\begin{table*}[htb]
	\centering
	\renewcommand\thetable{3}
	\scalebox{.93}{
		\begin{tabular}{lllllll}
			\toprule
			
			Method&En-Fr & Fr-En &En-Ro&Ro-En& En-Et & Et-En \\
			\midrule
			\citet{lample2017unsupervised}&15.05&14.31&n/a&n/a&n/a&n/a\\
			\citet{DBLP:journals/corr/abs-1710-11041}&15.13&15.56&n/a&n/a&n/a&n/a\\
			\citet{lample2018phrase}&27.60&27.68&25.13&23.90&n/a&n/a\\
			\citet{DBLP:journals/corr/abs-1901-07291}&33.40&33.30&33.30&31.80&n/a&n/a\\
			\midrule
			UNMT&31.01 &31.06&33.63&31.89&14.89&20.61\\
			\;\;\;\;+ST-UT   &34.43++&33.56++&35.04++&32.94++& 17.05++ & 22.60++  \\
			\;\;\;\;+ST-PT &\textbf{35.58++}&\textbf{34.91++}&\textbf{35.96++}&\textbf{33.64++}&\textbf{17.97++}& \textbf{24.97++}  \\
			
			\bottomrule
			
	\end{tabular}}
	\caption{ Performance (BLEU score) of UNMT on the unbalanced training data scenario. Note that only 2 million Fr monolingual training data were used for En--Fr.
	The quantity ratio $\epsilon$ was set to 10\%. The number of epochs was set to two for both proposed strategies.  ``++" after a score indicates that the strategy was significantly better than the baseline at significance level $p<$0.01.}
	\label{tab:main_results}
\end{table*}

\section{Experiments}
\label{Experiments}
\subsection{Datasets}
We considered three language pairs in our simulation experiments: Fr--En, Romanian (Ro)--En and Estonian (Et)--En translation tasks. The statistics of the data are  presented in Table \ref{Tab:Statistics}. We used the monolingual WMT news crawl datasets\footnote{\url{http://data.statmt.org/news-crawl/}} for each language. For the high-resource languages En and Fr, we randomly extracted 50M sentences. For the low-resource languages Ro and Et, we used all available  monolingual news crawl training data. 
To make our experiments comparable with previous work \cite{DBLP:journals/corr/abs-1901-07291}, we report the results on newstest2014 for Fr--En, newstest2016 for Ro--En, and newstest2018 for Et--En.
\begin{table}[h]
	\center
	\renewcommand\thetable{2}
	\scalebox{.85}{
		\begin{tabular}{l|rr}
			\toprule
			
			Language & Sentences 	& Words           \\
			\midrule
			
			En & 50.00M&    1.15B                         \\	 
			Fr &  50.00M  &  1.19B                        \\
			Ro &  8.92M  &    207.07M                          \\
			Et &  3.00M  &  51.39M                           \\				
			
			\bottomrule
	\end{tabular}}
	\caption{Statistics of the monolingual corpora.}
	\label{Tab:Statistics}
\end{table}

For preprocessing, we used the \texttt{Moses} tokenizer \cite{koehn-etal-2007-moses}. 
To clean the data, we only applied the \texttt{Moses} script \texttt{clean-corpus-n.perl} to remove lines from the monolingual data containing more than 50 words. 
We  used a shared vocabulary for all  language pairs, with 60,000 subword tokens based on BPE \cite{sennrich2015neural}.

\subsection{UNMT Settings}
We used a transformer-based \texttt{XLM} toolkit
and followed the settings of \citet{DBLP:journals/corr/abs-1901-07291}
for UNMT: six layers for the encoder and the decoder. 
The dimensions of the hidden layers were set to 1024. The batch size was set to 2000 tokens. 
The Adam optimizer \cite{kingma2014adam} was used to optimize the model parameters. The initial learning rate  was 0.0001,  $\beta_1 = 0.9$, and $\beta_2 = 0.98$. 
We trained a specific cross-lingual  language model for each different training dataset. The  language model was used to initialize the full parameters of  the UNMT system.
Eight V100 GPUs were used to train all UNMT models.
We used the case-sensitive 4-gram BLEU score computed by the $multi-bleu.perl$ script from Moses \cite{koehn-etal-2007-moses} to evaluate the test sets.

\subsection{Main Results}
Table \ref{tab:main_results} presents the detailed BLEU scores of the UNMT systems on the En--Fr, En--Ro, and En--Et test sets. 
Our re-implemented baseline performed similarly to the state-of-the-art method of  \citet{DBLP:journals/corr/abs-1901-07291} on the En--Ro language pair. In particular,  we  used only 2 million Fr monolingual training data on the En--Fr language pair, so the re-implemented baseline performed slightly worse than \citet{DBLP:journals/corr/abs-1901-07291}.

Our proposed self-training mechanism  substantially outperformed  the corresponding baseline in all language pairs by 2--4 BLEU points. Regarding the two proposed strategies, the ST-PT strategy performed better than the ST-UT strategy by 1 BLEU point because the synthetic data are more directly integrated into  the training. For ST-UT, the synthetic data was just used as the target part. In contrast, the synthetic data was used as the source and target part for ST-PT. The synthetic parallel data could improve translation performance.
These results demonstrate that synthetic data  improve translation performance in our proposed self-training mechanism.  The detailed analyses of the hyper-parameters such as  quantity ratio $\epsilon$ and epoch number $k_1,k_2$ are provided in Appendix.

\begin{table*}[htb]
	\centering
	\scalebox{.88}{
		\begin{tabular}{ll}
			\toprule
		 Input& Ma pole oma loomingust kunagi kaugel ja {\color{red}tööd ma ei karda} .\\

			Reference& I 'm never far from my work and {\color{red}I 'm not afraid of work} .\\
			
			\midrule
			Baseline & I am never far from my work and {\color{red}work I 'm not afraid of} .\\
			+ST-PT& I 'm never far from my work and {\color{red}I 'm not afraid of the job} .\\
	
			\midrule
			\midrule
		 Input& Salvador Adame kadus {\color{red}läänepoolses Michoacani} osariigis kolm päeva pärast Valdezi tapmmist .\\
			
			Reference& 	 Salvador Adame disappeared {\color{red}in the western state of Michoacan} three days after Valdez was killed .\\
			
			\midrule
			Baseline& Salvador Adame disappeared {\color{red}in west Michoacan , Mexico} , three days after Valdezi was killed .\\
			
			+ST-PT& Salvador Adame disappeared {\color{red}in the western state of Michoacan} three days after Valdezi was killed .\\

			\bottomrule
			
	\end{tabular}}
	\caption{Comparison of translation results of baseline and +ST-PT system on the Et-En dataset. }
	\label{tab:case}
\end{table*}	
\subsection{Case Study}
Moreover, we analyze translation examples to further analyze the effectiveness of our proposed  self-training mechanism.
Table \ref{tab:case} shows two translation examples, which were generated by UNMT baseline system and +ST-PT system on the Et-En dataset, respectively. For the first example, +ST-PT method could make the translation more fluent, compared with the baseline system. For the second example, +ST-PT method could make the translation more accurate. These examples indicate that  our proposed self-training mechanism could be widely applied to  the unbalanced training data scenario.

\section{Conclusion}
UNMT has achieved remarkable results on massive monolingual corpora.
  However, a UNMT system usually does not perform well in a scenario where there is not an  adequate training corpus for one language.  Based on this unbalanced training data scenario, we proposed two self-training strategies for UNMT. Experimental results on several language pairs show that our proposed strategies substantially outperform UNMT baseline. 
 
\section*{Acknowledgments}
We are grateful to the anonymous reviewers and the area chair for their insightful comments and suggestions.
Tiejun Zhao was partially supported by  National Key Research and Development Program of China via grant 2020AAA0108001.
Masao Utiyama and Eiichiro Sumita were partly supported by the commissioned research program ``Research and Development of Advanced Multilingual Translation Technology” in the ``R\&D Project for Information and Communications Technology (JPMI00316)” of the Ministry of Internal Affairs and Communications (MIC), Japan.

  \bibliography{custom}

\begin{thebibliography}{20}
\expandafter\ifx\csname natexlab\endcsname\relax\def\natexlab#1{#1}\fi

\bibitem[{Artetxe et~al.(2018)Artetxe, Labaka, Agirre, and
  Cho}]{DBLP:journals/corr/abs-1710-11041}
Mikel Artetxe, Gorka Labaka, Eneko Agirre, and Kyunghyun Cho. 2018.
\newblock Unsupervised neural machine translation.
\newblock In \emph{Proceedings of the Sixth International Conference on
  Learning Representations}, Vancouver, Canada.

\bibitem[{He et~al.(2020)He, Gu, Shen, and Ranzato}]{he2019revisiting}
Junxian He, Jiatao Gu, Jiajun Shen, and Marc'Aurelio Ranzato. 2020.
\newblock Revisiting self-training for neural sequence generation.
\newblock In \emph{Proceedings of the 8th International Conference on Learning
  Representations}, Addis Ababa, Ethiopia.

\bibitem[{Kingma and Ba(2015)}]{kingma2014adam}
Diederik~P Kingma and Jimmy Ba. 2015.
\newblock Adam: A method for stochastic optimization.
\newblock In \emph{Proceedings of the Third International Conference on
  Learning Representations}, San Diego, California, {USA}.

\bibitem[{Koehn et~al.(2007)Koehn, Hoang, Birch, Callison-Burch, Federico,
  Bertoldi, Cowan, Shen, Moran, Zens, Dyer, Bojar, Constantin, and
  Herbst}]{koehn-etal-2007-moses}
Philipp Koehn, Hieu Hoang, Alexandra Birch, Chris Callison-Burch, Marcello
  Federico, Nicola Bertoldi, Brooke Cowan, Wade Shen, Christine Moran, Richard
  Zens, Chris Dyer, Ond{\v{r}}ej Bojar, Alexandra Constantin, and Evan Herbst.
  2007.
\newblock {M}oses: Open source toolkit for statistical machine translation.
\newblock In \emph{Proceedings of the 45th Annual Meeting of the Association
  for Computational Linguistics Companion Volume Proceedings of the Demo and
  Poster Sessions}, pages 177--180, Prague, Czech Republic.

\bibitem[{Lample and Conneau(2019)}]{DBLP:journals/corr/abs-1901-07291}
Guillaume Lample and Alexis Conneau. 2019.
\newblock Cross-lingual language model pretraining.
\newblock \emph{CoRR}, abs/1901.07291.

\bibitem[{Lample et~al.(2018{\natexlab{a}})Lample, Conneau, Denoyer, and
  Ranzato}]{lample2017unsupervised}
Guillaume Lample, Alexis Conneau, Ludovic Denoyer, and Marc'Aurelio Ranzato.
  2018{\natexlab{a}}.
\newblock Unsupervised machine translation using monolingual corpora only.
\newblock In \emph{Proceedings of the Sixth International Conference on
  Learning Representations}, Vancouver, Canada.

\bibitem[{Lample et~al.(2018{\natexlab{b}})Lample, Ott, Conneau, Denoyer, and
  Ranzato}]{lample2018phrase}
Guillaume Lample, Myle Ott, Alexis Conneau, Ludovic Denoyer, and Marc'Aurelio
  Ranzato. 2018{\natexlab{b}}.
\newblock Phrase-based {\&} neural unsupervised machine translation.
\newblock In \emph{Proceedings of the 2018 Conference on Empirical Methods in
  Natural Language Processing}, pages 5039--5049, Brussels, Belgium.

\bibitem[{McClosky et~al.(2006)McClosky, Charniak, and
  Johnson}]{mcclosky-etal-2006-effective}
David McClosky, Eugene Charniak, and Mark Johnson. 2006.
\newblock Effective self-training for parsing.
\newblock In \emph{Proceedings of the Human Language Technology Conference of
  the {NAACL}, Main Conference}, pages 152--159, New York City, {USA}.

\bibitem[{Scudder(1965)}]{DBLP:journals/tit/Scudder65a}
H.~J. Scudder. 1965.
\newblock Probability of error of some adaptive pattern-recognition machines.
\newblock \emph{{IEEE} Trans. Information Theory}, 11(3):363--371.

\bibitem[{Sennrich et~al.(2016{\natexlab{a}})Sennrich, Haddow, and
  Birch}]{P16-1009}
Rico Sennrich, Barry Haddow, and Alexandra Birch. 2016{\natexlab{a}}.
\newblock Improving neural machine translation models with monolingual data.
\newblock In \emph{Proceedings of the 54th Annual Meeting of the Association
  for Computational Linguistics (Volume 1: Long Papers)}, pages 86--96, Berlin,
  Germany.

\bibitem[{Sennrich et~al.(2016{\natexlab{b}})Sennrich, Haddow, and
  Birch}]{sennrich2015neural}
Rico Sennrich, Barry Haddow, and Alexandra Birch. 2016{\natexlab{b}}.
\newblock Neural machine translation of rare words with subword units.
\newblock In \emph{Proceedings of the 54th Annual Meeting of the Association
  for Computational Linguistics (Volume 1: Long Papers)}, pages 1715--1725,
  Berlin, Germany.

\bibitem[{Song et~al.(2019)Song, Tan, Qin, Lu, and Liu}]{song2019mass}
Kaitao Song, Xu~Tan, Tao Qin, Jianfeng Lu, and Tie-Yan Liu. 2019.
\newblock Mass: Masked sequence to sequence pre-training for language
  generation.
\newblock In \emph{Proceedings of the 36th International Conference on Machine
  Learning}, pages 5926--5936, Long Beach, California, {USA}.

\bibitem[{Sun et~al.(2019)Sun, Wang, Chen, Utiyama, Sumita, and
  Zhao}]{sun-etal-2019-unsupervised}
Haipeng Sun, Rui Wang, Kehai Chen, Masao Utiyama, Eiichiro Sumita, and Tiejun
  Zhao. 2019.
\newblock Unsupervised bilingual word embedding agreement for unsupervised
  neural machine translation.
\newblock In \emph{Proceedings of the 57th Annual Meeting of the Association
  for Computational Linguistics}, pages 1235--1245, Florence, Italy.

\bibitem[{Sun et~al.(2020{\natexlab{a}})Sun, Wang, Chen, Utiyama, Sumita, and
  Zhao}]{sun-etal-2020-knowledge}
Haipeng Sun, Rui Wang, Kehai Chen, Masao Utiyama, Eiichiro Sumita, and Tiejun
  Zhao. 2020{\natexlab{a}}.
\newblock Knowledge distillation for multilingual unsupervised neural machine
  translation.
\newblock In \emph{Proceedings of the 58th Annual Meeting of the Association
  for Computational Linguistics}, pages 3525--3535, Online.

\bibitem[{Sun et~al.(2020{\natexlab{b}})Sun, Wang, Chen, Utiyama, Sumita, and
  Zhao}]{9043536}
Haipeng Sun, Rui Wang, Kehai Chen, Masao Utiyama, Eiichiro Sumita, and Tiejun
  Zhao. 2020{\natexlab{b}}.
\newblock Unsupervised neural machine translation with cross-lingual language
  representation agreement.
\newblock \emph{IEEE/ACM Transactions on Audio, Speech, and Language
  Processing}, 28:1170--1182.

\bibitem[{Vincent et~al.(2010)Vincent, Larochelle, Lajoie, Bengio, and
  Manzagol}]{DBLP:journals/jmlr/VincentLLBM10}
Pascal Vincent, Hugo Larochelle, Isabelle Lajoie, Yoshua Bengio, and
  Pierre{-}Antoine Manzagol. 2010.
\newblock Stacked denoising autoencoders: Learning useful representations in a
  deep network with a local denoising criterion.
\newblock \emph{Journal of Machine Learning Research}, 11:3371--3408.

\bibitem[{Wu et~al.(2019)Wu, Wang, and Wang}]{wu-etal-2019-extract}
Jiawei Wu, Xin Wang, and William~Yang Wang. 2019.
\newblock Extract and edit: An alternative to back-translation for unsupervised
  neural machine translation.
\newblock In \emph{Proceedings of the 2019 Conference of the North {A}merican
  Chapter of the Association for Computational Linguistics: Human Language
  Technologies, Volume 1 (Long and Short Papers)}, pages 1173--1183,
  Minneapolis, Minnesota, {USA}.

\bibitem[{Yang et~al.(2018)Yang, Chen, Wang, and Xu}]{P18-1005}
Zhen Yang, Wei Chen, Feng Wang, and Bo~Xu. 2018.
\newblock Unsupervised neural machine translation with weight sharing.
\newblock In \emph{Proceedings of the 56th Annual Meeting of the Association
  for Computational Linguistics (Volume 1: Long Papers)}, pages 46--55,
  Melbourne, Australia.

\bibitem[{Yarowsky(1995)}]{yarowsky-1995-unsupervised}
David Yarowsky. 1995.
\newblock Unsupervised word sense disambiguation rivaling supervised methods.
\newblock In \emph{33rd Annual Meeting of the Association for Computational
  Linguistics}, pages 189--196, Cambridge, Massachusetts, USA.

\bibitem[{Zhang and Zong(2016)}]{zhang-zong-2016-exploiting}
Jiajun Zhang and Chengqing Zong. 2016.
\newblock Exploiting source-side monolingual data in neural machine
  translation.
\newblock In \emph{Proceedings of the 2016 Conference on Empirical Methods in
  Natural Language Processing}, pages 1535--1545, Austin, Texas, {USA}.

\end{thebibliography}
\bibliographystyle{acl_natbib}
\newpage
\appendix

\section{Appendix}
\label{sec:appendix}
\subsection{Quantity Ratio Analysis}
\begin{figure}[ht]
  \centering

	\setlength{\abovecaptionskip}{0pt}
	
	\begin{center}
		\pgfplotsset{width=6.5cm,compat=1.14,every axis/.append style={thick}}
		\begin{tikzpicture}
		\tikzset{every node}=[font=\small]
		\begin{axis}
		[enlargelimits=0.13, tick align=outside, legend style={cells={anchor=west},legend pos=south east, legend columns=1,every axis legend/.append style={
				at={(1.4,0)}}}, xticklabels={ $1\%$,$5\%$,$10\%$, $30\%$,$50\%$,$100\%$},
		xtick={0,1,2,3,4,5},
		axis y line*=left,
		axis x line*=left,
		ylabel={BLEU score},xlabel={epoch size},font=\small]
		\addplot+ [sharp plot, mark=triangle*,mark size=1.2pt,mark options={solid,mark color=lightgray}, color=lightgray] coordinates
		{ (0,31.01)(1,31.01)(2,31.01)(3,31.01)(4,31.01)(5,31.01)};
		\addlegendentry{\tiny En-Fr-baseline}
		
		\addplot+ [sharp plot,mark=square*,mark size=1.2pt,mark options={solid,mark color=cyan}, color=cyan] coordinates
		{ (0,31.51)(1,32.7)(2,33.66)(3,33.69)(4,33.65)(5,33.68)  };
		\addlegendentry{\tiny En-Fr-ST-UT}

		
		\addplot+ [sharp plot, mark=*,mark size=1.2pt,mark options={solid,mark color=orange}, color=orange] coordinates
		{ (0,31.61)(1,33.41)(2,34.54)(3,34.50)(4,34.57)(5,34.55)};
		\addlegendentry{\tiny En-Fr-ST-PT}

		\end{axis}
		\end{tikzpicture}

	\end{center}

  		\caption{\label{fig:ratio} Effect of the quantity ratio $\epsilon$  on UNMT performance for  the En--Fr translation tasks.}
  \end{figure}
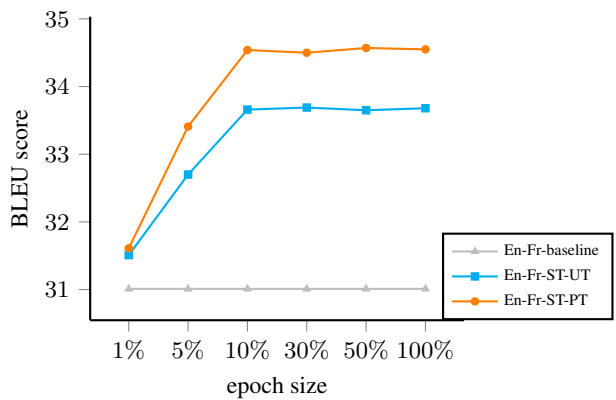

We  investigated the effect of  quantity ratio $\epsilon$  on UNMT performance for  the En--Fr translation task during the first epoch of our proposed self-training methods. As  shown in Fig. \ref{fig:ratio}, $\epsilon$ ranging from 1\% to 100\%  all enhanced UNMT performance and  the performance was similar when the quantity ratio $\epsilon$ was greater than 10\%.
The UNMT model converged faster with less data. Therefore, we selected 10\% as the quantity ratio $\epsilon$ for our proposed self-training methods.

\subsection{Epoch Number Analysis}

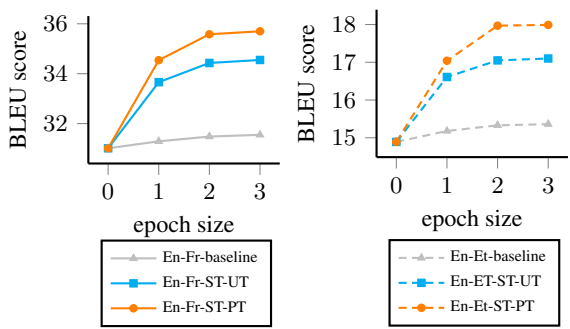
\begin{figure}[ht]
  \centering
\begin{minipage}{.48\linewidth}
	\setlength{\abovecaptionskip}{0pt}
	
	\begin{center}
		\pgfplotsset{width=4.1cm,compat=1.14,every axis/.append style={thick}}
		\begin{tikzpicture}
		\tikzset{every node}=[font=\small]
		\begin{axis}
		[enlargelimits=0.13, tick align=outside, legend style={cells={anchor=west},legend pos=south east, legend columns=1,every axis legend/.append style={
				at={(0.96,-1.11)}}}, xticklabels={ $0$, $1$,$2$, $3$},
		xtick={0,1,2,3},
		axis y line*=left,
		axis x line*=left,
		ylabel={BLEU score},xlabel={epoch size},font=\small]
		\addplot+ [sharp plot, mark=triangle*,mark size=1.2pt,mark options={solid,mark color=lightgray}, color=lightgray] coordinates
		{ (0,31.01)(1,31.29)(2,31.48)(3,31.55)};
		\addlegendentry{\tiny En-Fr-baseline}
		
		\addplot+ [sharp plot,mark=square*,mark size=1.2pt,mark options={solid,mark color=cyan}, color=cyan] coordinates
		{ (0,31.01)(1,33.66)(2,34.43)(3,34.55) };
		\addlegendentry{\tiny En-Fr-ST-UT}

		
		\addplot+ [sharp plot, mark=*,mark size=1.2pt,mark options={solid,mark color=orange}, color=orange] coordinates
		{ (0,31.01)(1,34.54)(2,35.58)(3,35.70)};
		\addlegendentry{\tiny En-Fr-ST-PT}

		\end{axis}
		\end{tikzpicture}

	\end{center}

  \end{minipage}
\begin{minipage}{.48\linewidth}	\setlength{\abovecaptionskip}{0pt}
	
	\begin{center}
		\pgfplotsset{width=4.1cm,compat=1.14,every axis/.append style={thick}}
		\begin{tikzpicture}
		\tikzset{every node}=[font=\small]
		\begin{axis}
		[enlargelimits=0.13, tick align=outside, legend style={cells={anchor=west},legend pos=south east, legend columns=1,every axis legend/.append style={
				at={(0.96,-1.15)}}}, xticklabels={ $0$, $1$,$2$, $3$},
		xtick={0,1,2,3},
		axis y line*=left,
		axis x line*=left,ylabel={BLEU score},
		xlabel={epoch size},font=\small]

\addplot+ [sharp plot,densely dashed, mark=triangle*,mark size=1.2pt,mark options={solid,mark color=lightgray}, color=lightgray] coordinates
		{ (0,14.89)(1,15.18)(2,15.33)(3,15.36)};
		\addlegendentry{\tiny En-Et-baseline}		
				\addplot+ [sharp plot,densely dashed,mark=square*,mark size=1.2pt,mark options={solid,mark color=cyan}, color=cyan] coordinates
		{ (0,14.89)(1,16.61)(2,17.05)(3,17.10) };
		\addlegendentry{\tiny En-ET-ST-UT}

				\addplot+ [sharp plot,densely dashed, mark=*,mark size=1.2pt,mark options={solid,mark color=orange}, color=orange] coordinates
		{ (0,14.89)(1,17.04)(2,17.97)(3,17.99)};
		\addlegendentry{\tiny En-Et-ST-PT}
		\end{axis}
		\end{tikzpicture}

	\end{center}

  \end{minipage}
  		\caption{\label{fig:epoch} Effect of the number of epochs  on UNMT performance for  the En--Fr and En--Et translation tasks.}
  \end{figure}

In Figure \ref{fig:epoch}, we empirically demonstrate how the number of epochs affects the UNMT performance on the En--Fr and En--Et translation tasks.
We found that the use of additional epochs has little influence on the baseline system.  In contrast, increasing the number of epochs for our proposed strategies can improve performance because the quality of the synthetic data used by the UNMT model is better after more epochs; however, the improvement decreases as additional epochs are added. Considering the computational cost of synthetic data generation, we trained the UNMT model for only two epochs.

\end{document}